# Connecting Fairness in Machine Learning with Public Health Equity

Shaina Raza
*University of Toronto*
Toronto, Canada
shaina.raza@utoronto.ca

*Abstract*— Machine learning (ML) has become a critical tool in public health, offering the potential to improve population health, diagnosis, treatment selection, and health system efficiency. However, biases in data and model design can result in disparities for certain protected groups and amplify existing inequalities in healthcare. To address this challenge, this study summarizes seminal literature on ML fairness and presents a framework for identifying and mitigating biases in the data and model. The framework provides guidance on incorporating fairness into different stages of the typical ML pipeline, such as data processing, model design, deployment, and evaluation. To illustrate the impact of biases in data on ML models, we present examples that demonstrate how systematic biases can be amplified through model predictions. These case studies suggest how the framework can be used to prevent these biases and highlight the need for fair and equitable ML models in public health. This work aims to inform and guide the use of ML in public health towards a more ethical and equitable outcome for all populations.

## I. INTRODUCTION

Health equity [1] is a crucial principle in public health, which aims to eliminate disparities in health outcomes and healthcare access among various populations. The World Health Organization (WHO) [2] and the United Nations (UN) [3] both prioritize health equity as a key aspect of their missions to improve global health outcomes. However, despite these efforts, disparities in health outcomes continue to persist, particularly among marginalized and underserved populations [4].

Machine learning (ML) has the potential to transform the way we approach health and healthcare with its advanced analytical and predictive capabilities. ML can aid in comprehending complex health systems, identifying disease patterns and trends, and improving patient outcomes [5]. However, it is essential to exercise caution when employing ML and consider potential biases and inequalities that may be present in the data used to train these models [6], [7]. Such biases can lead to discrimination and unjust outcomes for specific populations, exacerbating existing health disparities [8].

This study aims to promote health equity through ML by reviewing the literature on ML fairness and presenting a novel ML pipeline approach to integrate fairness into various stages of a standard ML pipeline. Although fair ML has been explored in artificial intelligence (AI) literature [9], [10], its implementation in public health has received limited attention. This study endeavors to provide the public health community with accessible methods for ensuring equitable outcomes when using ML. The specific contributions of this research are:

- Summarizing the concepts of fair ML and presenting an ML pipeline approach for public health use to achieve equitable outcomes.
- Providing examples that demonstrate the importance of the pipeline approach and how disparities can be amplified and mitigated through ML.
- Offering straightforward and accessible methods for the public health community to incorporate fairness in their use of ML.

Unlike previous studies [6], [11]–[13] that have primarily focused on specific applications of ML in healthcare, this review adopts a different approach by presenting a methodology for incorporating fairness during various stages of a standard ML pipeline. This pipeline idea is proposed based on a thorough review of pertinent literature on fair ML. The study primary focus is on promoting health equity for marginalized and underserved populations and providing straightforward and accessible methods for the public health community.

## II. BACKGROUND

Health equity [1], [14] is a fundamental principle in public health that emphasizes the importance of ensuring equal access to healthcare resources, opportunities, and outcomes for all individuals, regardless of their background or social status. It seeks to address the root causes of health disparities and strives to eliminate barriers that prevent certain populations from achieving optimal health.

Fairness in ML [9], [15], [16] refers to the development and application of ML models that minimize biases and disparities in their predictions, ensuring equitable outcomes for different groups in a population. As ML models are increasingly used in various domains, including healthcare, finance, and criminal justice, it is crucial to address potential biases in the data and algorithms used. Fairness in ML involves identifying and mitigating these biases to prevent discriminatory consequences and to promote equitable decision-making [17]. This subsection discusses previous studies that have investigated the use of ML in the field of health, with a focus on fairness and equity.

Rajkomar et al. [18] emphasize the importance of ensuring fairness in ML to advance health equity in healthcare. They highlight the need for implementing inclusive and equitable research guidelines and utilizing technical solutions to address biases in the models to prevent them from amplifying and contribute to promoting health equity.

Fletcher et al. [19] address the use of ML and artificial intelligence (AI) in Low- and Middle-Income Countries (LMICs). They suggest the use of three criteria, Appropriateness, Fairness, and Bias, to help evaluate the use of AI and ML in the global health context. Appropriateness involves deciding the appropriate use of the algorithm in the local context and matching the model to the target population. Bias refers to systematic tendencies in the model that may favor one demographic group over another. Fairness involves examining the impact on various demographic groups and choosing a mathematical definition of group fairness that satisfies cultural, legal, and ethical

Mhasawade et al. [20] discuss the potential for using ML in the field of public and population health. They highlight the importance of considering the connection between social, cultural, and environmental factors and their effect on health. The authors also emphasize the importance of addressing health equity and disparities through ML methods that focus on algorithmic fairness.

Thomasian et al. [7] discuss the impact of AI in the field of population and public health and the potential consequences of unmitigated bias in AI-driven health care. They argue that a consensus on the regulation of algorithmic bias at the policy level is necessary to ensure the ethical integration of AI into the health system. The authors present three overarching principles for mitigating bias in healthcare AI and call for a framework for federal oversight.

Wesson et al. [21] discuss the potential benefits and drawbacks of using big data in public health research. They caution against perpetuating discriminatory practices and highlight the importance of incorporating an equity lens in order to advance equity in public health. The authors frame the concept of a sixth V, virtuosity, in the big data context of the five Vs (volume, velocity, veracity, variety, and value). The idea is to encompass equity and justice frameworks and provide examples of analytic approaches to improving equity in big data.

Sikstrom et al. [22] conduct a comprehensive environmental scan of literature on fairness in the context of AI and ML. The main aim of their study is to advance efforts in operationalizing fairness in medicine by synthesizing a broad range of literature. The authors searched electronic databases and conducted hand searches to gather data from 213 selected publications, which were then analyzed using rapid framework analysis. The search and analysis were conducted in two rounds to explore both pre-identified and emerging issues related to fairness in ML.

Gervasi et al. [23] address the issue of fairness, equity, and bias in the use of ML algorithms in the health insurance industry. They provide a guide to the data ecosystem used by health insurers and highlight potential sources of bias in the machine learning pipelines.

Gichoya et al. [11] discuss the use of ML in healthcare and the need to establish guidelines and protocols for its development and implementation. They note that while technical considerations are well addressed in current guidelines, there is a lack of engagement with issues of fairness, bias, and unintended disparate impact.

In summary, these works discuss the rapid growth of ML for healthcare and the need to formalize processes and procedures for characterizing and evaluating the performance of these tools.

## III. FAIR MACHINE LEARNING FRAMEWORK

### A. Preliminaries

The definitions of key terms used in this work have been extracted from relevant literature sources [9], [17], [24] and given below.

- *Bias*: A systematic error in an ML model, data or even research.
- *Fairness*: The process of addressing biases present in data and algorithms to mitigate their adverse impact.
- *Protected Attributes:* Characteristics that divide a population into distinct groups (e.g. race, gender, age).
- *Privileged Values of Protected Attributes:* Characteristics that denote groups that have a systematic advantage (e.g., male gender, white race in many cases).
- *Underprivileged Groups of Protected Attributes*: Characteristics that denote groups that face prejudice.
- *Favorable Outcome:* A desirable result, such as obtaining a loan or insurance.
- *Demographic Parity*: A measure of fairness that requires that the distribution of positive predictions be equal for different protected attribute groups.
- *Disparate Impact (DI):* A metric used to evaluate fairness in ML models. DI compares the percentage of favorable outcomes for unprivileged groups to the percentage of favorable outcomes for privileged groups.
- *Equal Opportunity*: A measure of fairness that requires that the false positive rate be equal across different protected attribute groups.
- *Counterfactual Fairness*: A measure of fairness that requires that the outcomes of a model's predictions be fair for different counterfactual scenarios, e.g., scenarios where a person's protected attribute values are different.

### B. Pipeline Approach

Our framework for achieving fairness in ML is based on an extensive literature review. The framework, shown in Figure 1, is structured as a pipeline [25] approach, with specific steps outlined below to create fair and unbiased models.

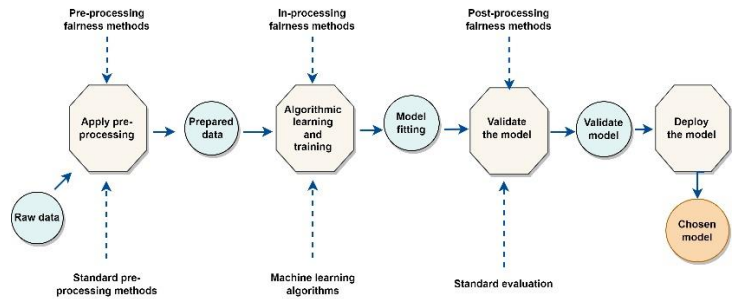

Fig. 1. Fair Machine Learning Pipeline

*Data Pre-processing:* This step involves preparing the data for the model by cleaning, normalizing, and transforming it, as well as splitting it into training, validation, and test sets. In this stage, fairness is achieved through pre-processing techniques such as editing feature values [26], learning fair representations [27], and re-sampling or re-weighting [28] techniques. The end result of this step is a dataset that is prepared for the next stage in the pipeline.

*Algorithmic Learning and Training*: This step involves selecting a suitable ML algorithm and training the model using the prepared data. In this stage, fairness is achieved through in-processing techniques such as fair classification, clustering, adversarial learning [29], and counterfactual fair learning [30] algorithms.

*Validate Model*: This step involves evaluating the performance of the model using various metrics such as accuracy, precision, recall, F1 score, etc. It also includes tuning the hyperparameters to improve the model performance. Fairness in this step, also known as post-processing fairness, is achieved by employing techniques such as counterfactual analysis, calibration, and interpretability techniques, which work by altering the predictions made by a model [31].

*Deploy:* This step involves deploying the final model in a production environment [25], where it can be used to make predictions on new data. It is important to note that monitoring and testing the model performance on an ongoing basis is crucial to ensure continued fairness and bias mitigation.

By following this pipeline approach, the proposed framework ensures that fairness is considered and integrated throughout the entire ML process, from data pre-processing to deployment. This holistic approach to fairness helps to mitigate biases and ensure the development of fair and unbiased models.

## IV. EXAMPLE USE CASES

In the field of precision medicine and public health, researchers have investigated potential racial disparities in various aspects, such as the administration of a genetic test for a specific type of cancer [32] and the impact of COVID-19 on different demographic groups [33]. The findings from these studies are both alarming and compelling, as they demonstrate that biases present in healthcare models can lead to systemic unfair treatment and adverse outcomes for underprivileged groups. If such biases remain unaddressed within the data, it is highly likely that they will be exacerbated through the predictions generated by ML models. This highlights the urgent need to develop and implement fair and unbiased ML algorithms to ensure equitable access to healthcare and improve outcomes for all individuals, regardless of their demographic background.

The first step in addressing biases in ML models is to identify protected attributes and their corresponding privileged and unprivileged groups. In the genetic test for cancer example [32], the authors analyzed a large electronic health record dataset and discovered that African American patients were significantly less likely to receive the genetic test compared to white and/or other race patients, despite having a higher incidence of the cancer the test was designed to detect. In the COVID-19 use case example [33], the study found that COVID-19 disproportionately affected communities of color, with African American, Hispanic, and Native American populations experiencing higher rates of infection, hospitalization, and death compared to white populations.

To address these issues, the Fair ML pipeline presented in this work can be utilized to ensure that the algorithms used for predictions and decisions are fair and unbiased. Fairness methods can be incorporated at specific stages of the ML pipeline, depending on the nature of the problem and the biases identified in the dataset.

During the data pre-processing stage, the dataset can be scrutinized to detect potential sources of bias and discrimination, and adjustments can be made to guarantee balanced and representative data of the population of interest. Fairness metrics, such as disparate impact [26] or equal opportunity [34], can be applied to measure fairness in the data.

In the algorithmic learning and training stage, the ML model can be trained on the transformed data (from the pre-processing stage) to ensure fairness and unbiasedness across different demographic groups. Techniques like regularization [34] or adversarial learning [35] can be employed to encourage the model to make equally accurate predictions for all individuals, and fairness metrics can be used to measure and correct bias during the training process.

In the validation stage, if biases are identified in the model's predictions, post-processing fairness techniques, such as equalized odds [36] or calibrated equalized odds [31] can be applied. These techniques, such as adjusting the decision threshold or predicted probabilities, ensure that the model's predictions are equally accurate for all individuals.

By incorporating fairness methods at relevant stages of the ML pipeline, the proposed framework can help address the biases identified in the precision medicine and COVID-19 outcome examples, ensuring that the resulting models are fair, equitable, and produce accurate, unbiased predictions for all individuals, irrespective of their race or ethnicity. This approach not only mitigates potential sources of bias in healthcare datasets but also contributes to improved health outcomes for marginalized and underrepresented populations, ultimately leading to a more just and equitable healthcare system for all.

## V. DISCUSSION

*A. Main Findings*

Fairness in ML plays a vital role in promoting public health equity by addressing biases in healthcare models and ensuring that predictive algorithms are both accurate and equitable. By connecting the principles of fair ML [9], [37] with the broader goals of public health equity, scientists and healthcare professionals can work together to develop innovative techniques and methodologies that address existing biases and disparities. This collaborative approach not only mitigates the potential for discriminatory outcomes but also contributes to improved health outcomes for all individuals, regardless of their demographic background.

Furthermore, by integrating interdisciplinary knowledge from fields such as computer science, statistics, and public health, researchers can develop robust fairness-aware ML algorithms that uncover hidden patterns, enabling a deeper understanding of the complex relationship between demographic factors and health outcomes. However, addressing public health equity requires a comprehensive approach that includes analyzing health disparities, targeting interventions, addressing social determinants of health [38], [39], and promoting equal access to healthcare services. By incorporating fair ML models into public health decision-making processes, we can ensure that predictions and decisions are fair and do not perpetuate or exacerbate existing health disparities. Ultimately, the connection between fair ML and public health equity is critical to creating a more just and equitable system for all.

*B. Practical Impact*

The practical impact of fairness in ML can be significant in several ways. Fairness helps prevent discrimination and bias in predictive models, ensuring equitable decision-making and reducing the risk of harmful actions based on flawed models [20]. This can lead to greater trust in AI systems. Moreover, fair ML can lead to better outcomes for individuals and society as a whole. In healthcare, fair predictive models can help identify high-risk patients, enabling targeted care and interventions, leading to improved patient outcomes and reduced healthcare costs [19], [40].

*C. Limitations*

The use of fairness in ML is subject to several limitations. There is a potential for over-correction or unintended consequences, introducing new biases or errors into the model. Additionally, there is difficulty in defining and measuring fairness, leading to different outcomes and interpretations. The lack of diverse and representative data is another challenge, potentially resulting in unfair or inaccurate models. Furthermore, the complexity of models makes interpretation and understanding challenging.

In the public health domain, additional limitations include small sample sizes and imbalanced datasets, making it difficult to train fair and accurate models. There is also the potential for confounding variables affecting accuracy and fairness. Finally, there are challenges in integrating predictive models into existing healthcare systems and workflows.

*D. Recommendations*

To address public health equity, a comprehensive approach is required. One important recommendation is to analyze health disparities among different population groups and target interventions accordingly. Another strategy is to address the social determinants of health, such as poverty, education, and access to healthcare. These underlying issues play a major role in health disparities, and addressing them can help reduce them.

Increasing access to quality healthcare for all, regardless of socioeconomic status, is also essential for promoting health equity. In addition, government policies that address issues such as poverty, education, and access to healthcare can help promote health equity. Engaging and empowering communities is crucial to ensure that interventions are tailored to their specific needs and are more likely to be successful.

Regularly monitoring and evaluating interventions is also essential for understanding their effectiveness and making any necessary adjustments to improve their impact. Finally, using fair ML models that align with the principles of distributive justice, such as equity, representation, and accountability, can help ensure that predictions and decisions made by the models are fair and do not perpetuate or exacerbate existing health disparities.

The connection between fair ML and public health equity is critical, as the use of unbiased ML models can aid in the reduction of existing health disparities and promote equal access to healthcare services. We can effectively analyse and address the underlying factors that contribute to health inequities in various population groups by ensuring that ML algorithms are fair and unbiased. Fair ML models can be used to identify at-risk populations, optimise resource allocation, and tailor public health interventions to specific community needs. Furthermore, incorporating fair ML into public health decision-making processes aids in the elimination of potential biases in predictive models, preventing the perpetuation or exacerbation of existing health disparities.

## VI. CONCLUSION

This paper has explored the importance of fairness in machine learning, providing a comprehensive framework for developing fair and accurate predictive models. Through the examination of various case studies in the public health domain, this paper has demonstrated the practical applications and potential impact of fair ML models on healthcare outcomes. By offering recommendations and addressing limitations, this paper contributes to the ongoing discourse on the responsible use of ML in public health and other domains, paving the way for future research and advancements in ethical AI.